\def\year{2022}\relax
\documentclass[letterpaper]{article} 
\usepackage{aaai22}  
\usepackage{times}  
\usepackage{helvet}  
\usepackage{courier}  
\usepackage[hyphens]{url}  
\usepackage{graphicx} 
\usepackage{color}
\usepackage{natbib}  
\usepackage{caption} 
\usepackage{booktabs} 
\DeclareCaptionStyle{ruled}{labelfont=normalfont,labelsep=colon,strut=off} 
\usepackage{algorithm}
\usepackage{algorithmic}
\usepackage{amsmath,amssymb,amsfonts,eucal}
\usepackage{relsize}
\usepackage{subfigure}
\usepackage{newfloat}
\usepackage{listings}
\floatstyle{ruled}
\newfloat{listing}{tb}{lst}{}
\floatname{listing}{Listing}
\title{Improving Multi-gerneration Robustness of Learned Image Compression}
\author{
    Litian Li,\textsuperscript{\rm 1}
    Zheng Yang,\textsuperscript{\rm 1}
    Ronggang Wang \textsuperscript{\rm 1 \rm 2}
}
\affiliations{

    \textsuperscript{\rm 1} Shenzhen Graduate School, Peking University, China \\
    \textsuperscript{\rm 2} Peng Cheng Laboratory, Shenzhen, China \\
    \{lilitian, zyang\}@stu.pku.edu.cn \{rgwang\}@pkusz.edu.cn
}

\usepackage{bibentry}

\begin{document}

\maketitle

\begin{abstract}
Benefit from flexible network designs and end-to-end joint optimization approach, learned image compression (LIC) has demonstrated excellent coding performance and practical feasibility in recent years. However, existing compression models suffer from serious multi-generation loss, which always occurs during image editing and transcoding. During the process of repeatedly encoding and decoding, the quality of the image will rapidly degrade, resulting in various types of distortion, which significantly limits the practical application of LIC. In this paper, a thorough analysis is carried out to determine the source of generative loss in successive image compression (SIC). We point out and solve the quantization drift problem that affects SIC, reversibility loss function as well as channel relaxation method are proposed to further reduce the generation loss. Experiments show that by using our proposed solutions, LIC can achieve comparable performance to the first compression of BPG even after 50 times reencoding without any change of the network structure.
\end{abstract}

\section{Introduction}\label{sec:introduction}
Image compression is one of the most fundamental technologies in the multimedia field. Efficient image compression technology, especially lossy compression technology, provides support for the storage and transmission of massive image data. Traditional coding standards, such as JPEG~\cite{wallace1992jpeg}, BPG~\cite{BPG}, and the latest VVC~\cite{VVC}, rely on hand-crafted modules to remove spatial and statistical redundancy in image data to achieve the purpose of compression.

In recent years, image coding methods based on end-to-end optimization have been rapidly explored and developed, and show promise to become the next-generation coding standard. On the one hand, with the powerful image understanding and generation capabilities of deep learning, some very recent works outperform VVC in PSNR and MS-SSIM ~\cite{Gao_2021_ICCV, guo2021causal, xie2021enhanced, chen_two-stage_2022, he2022elic}. On the other hand, to meet the needs of industrial applications, researchers have designed flexible modules to implement variable bitrate~\cite{cui2021asymmetric} and scalable coding~\cite{guo2019deep}. For HDR images~\cite{Cao_Jiang_Li_Wu_Ye_2022}, stereo images~\cite{deng2021deep}, and omnidirectional images~\cite{li2021pseudocylindrical}, learning-based methods also showed superiority over traditional codecs. Despite the remarkable progress of deep learning in the image compression field, there are two problems that need to be solved before coming into our lives. One is the model lightweight, mainly to reduce the network complexity, which is crucial for mobile devices~\cite{nongpu}; another is the model robustness, which aims to make the model more reliable. In this paper, we focus on improving the stability of the models during successive compression, called multi-generation robustness~\cite{RobustJPEG}.

\begin{figure*}[htbp]
    \centering
        \subfigure[``kodim24.png'' after 50 times reencoding]{
        \includegraphics[width=0.6\linewidth]{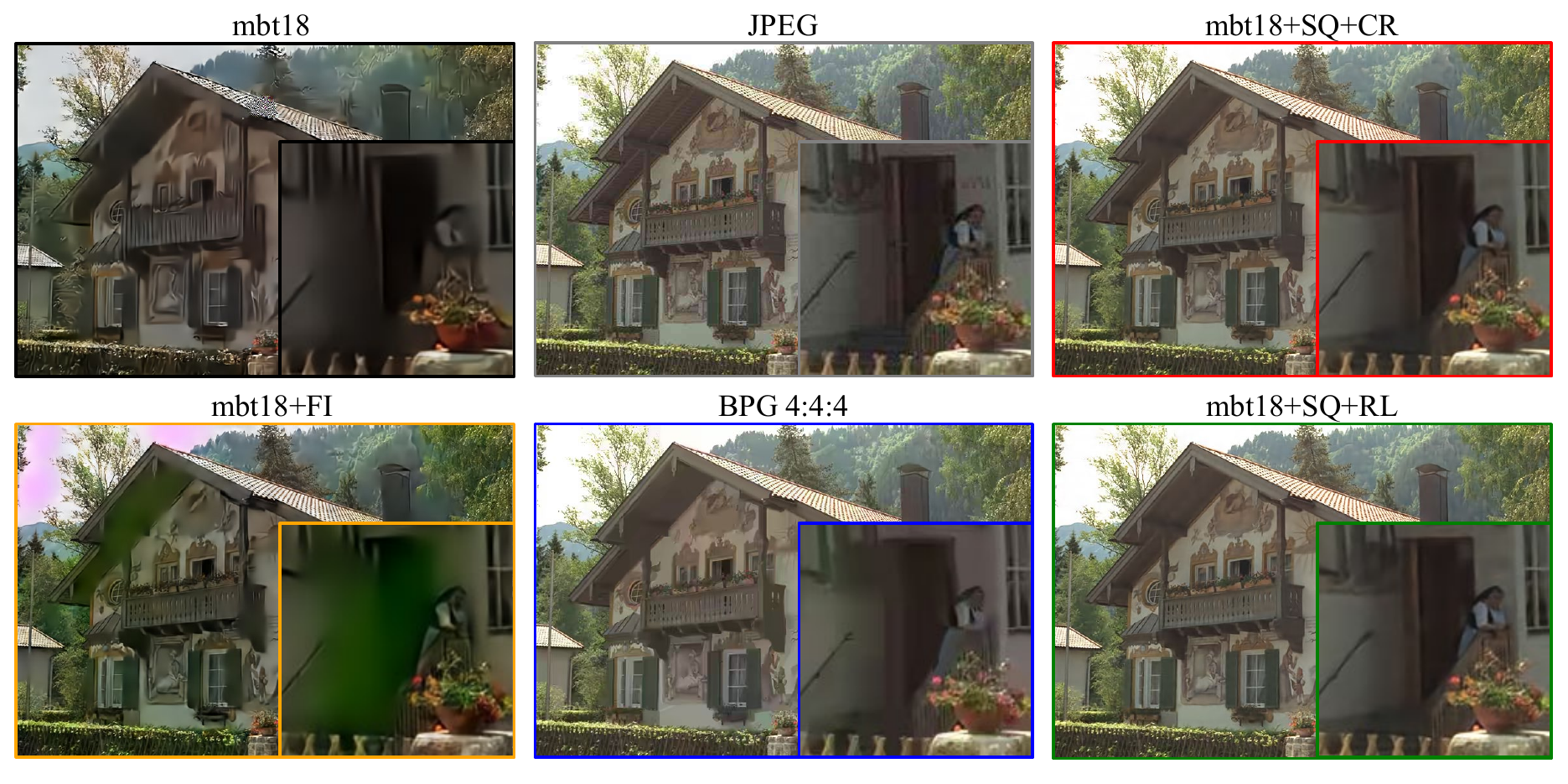}
    }
    \centering
     \subfigure[RD performance during SIC]{
        \includegraphics[width=0.37\linewidth]{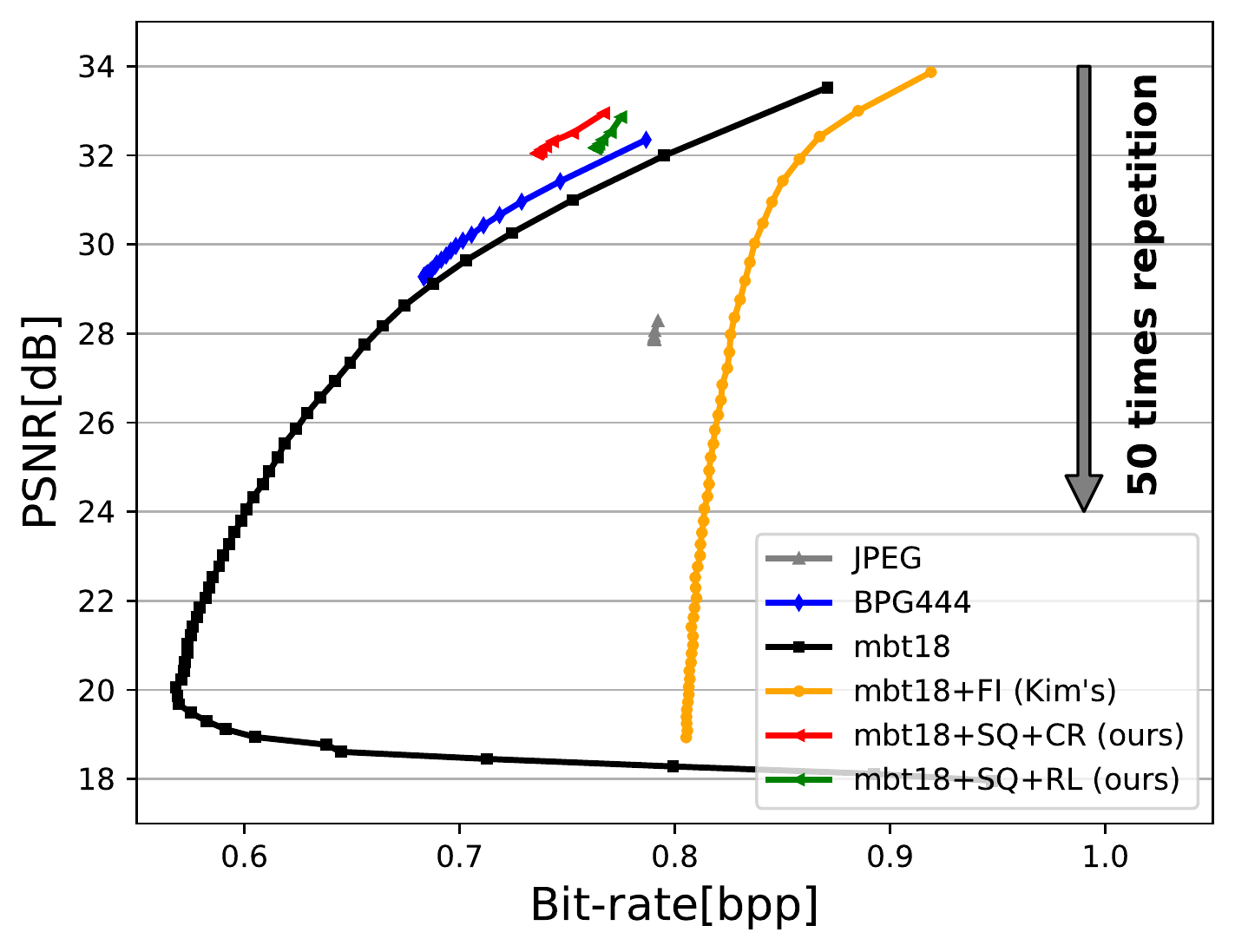}
    }
    \caption{The quality loss and the bitrate change after each recompression cycle. ``mbt18''~\cite{minnen2018joint} is the baseline model, ``FI'' represents feature identity loss proposed in~\cite{2020Instability}. Enhanced by our proposed solutions, the image quality degradation stops quickly and maintain excellent perceptual quality and RD performance during SIC.}
    \label{Figure1}
\end{figure*}

Multi-generation coding is a repeated compression and decompression process of images or videos~\cite{RobustMPEG}, which often occurs in multimedia application scenarios such as editing, transcoding, and redistribution. In the case of lossless compression, the repeatedly encoded and decoded image is identical to the original image. For lossy compression methods, the first compression will induce distortion, but the distortion will ideally not increase when the decoded image is reencoded with the same settings. However, the generation robustness will be impacted by many factors, which have been extensively studied in traditional coding standards~\cite{RobustJPEG, RobustH264, RobustHEVC}. The instability of successive deep image compression is also discussed in previous work~\cite{2020Instability}. After an image is repeatedly compressed by an existing compression model, the quality will drastically degrade, resulting in blurring and color casts. Moreover, compression artifacts can accumulate in multiple decompression-re-compression cycles and eventually corrupt the image. The salient patterns not only degrade the visual experience, but the induced high-frequency noise can significantly increase the bitrate of the image, which is particularly detrimental to practical applications.

This paper aims to thoroughly analyze the factors that affect the robustness of LIC for multiple generations and to enhance the stability of SIC without impacting the rate-distortion performance of the first encoding. According to our analysis, the quantization strategy, the reversibility of transformation, and the adaptability of the network play key roles in multi-generation robustness. To reduce error accumulation in SIC cycles, we suggest using straight quantization (SQ) instead of corrected quantization. To increase the reversibility of transformation, reversibility loss (RL) and channel relaxation (CR) methods are proposed. Experiments show that the robustness of the model enhanced by our solutions significantly outperforms previous work~\cite{2020Instability}, and our methods can be easily extended to different learned compression models. Figure~\ref{Figure1} shows the extraordinary performance of our two solutions. The contributions of this paper are summarized as follows:

\begin{itemize}
    \item For the first time, we have discovered that the wildly used corrected quantization causes quantization drift in SIC cycles, resulting in continuous degradation of image quality. Instead, using straight quantization can minimize generation loss without impacting the performance of the first compression.
    \item The effect of network adaptability and reversibility on successive compression is thoroughly analyzed in this paper. We report that invertible neural networks are fragile in the SIC task, which is similar to the result on image rescaling~\cite{pan2022towards}. To improve the reversibility of existing variational models, we propose reversibility loss function and channel relaxation strategy. To our best knowledge, we are the first to investigate the effect of channel number of latent representations on multi-generation robustness.
    \item By using our simple yet effective methods, existing models can achieve better multi-generation robustness than BPG~\cite{BPG} without changing the network structure. Even after 50 re-compression cycles, rate-distortion performance is still comparable to BPG, which provides further promise for the practical application of LIC.
\end{itemize}


\section{Related Works}\label{sec:relatework}
\subsection{Learned Variational Image Compression}
Recent lossy image compression frameworks are based on transform coding~\cite{goyal2001theoretical}, where the encoder applies an analysis transform $g_a$ mapping the input image $x$ to its latent representation $y$. The latent representation $y$ is quantized by $Q(\cdot)$ and entropy coded. For reconstruction, compressed latent $\hat y$ obtained from entropy decoding is passed through synthesis transform $g_s$ to yield $\hat x$. Mathematically, we can write:
\begin{equation}
    y = g_a(x); \quad
    \hat y = Q(y); \quad
    \hat x = g_s(\hat y).
\end{equation}

To deal with the zero gradient problem caused by quantization, additive uniform noise is applied to $y$ as a continuous approximation during training with $\hat y = y + U(-0.5, 0.5)$ ~\cite{balle2018variational}. Then this approach is equivalent to a variational autoencoder~\cite{Kingma2014}. The framework is optimized in an end-to-end manner:
\begin{equation}
    \mathcal{L} = R(\hat y) + \lambda D(x, \hat x),
    \label{equation:RD}
\end{equation}
where the hyper-parameter $\lambda$ is used to realize the trade-off between the estimated bitrate $R$ and image reconstruction distortion $D$.

A key challenge in LIC is estimating the entropy of quantized latent $\hat y$, for more accurate models typically result in better performance. As shown in Figure~\ref{Figure2}, existing models usually adopt a joint hyperprior and context entropy model to obtain a more accurate probability estimation. Each symbol $y_i$ of the latent representation is usually modeled as a Gaussian distribution $N(\mu_i,\sigma_i^2)$. The hyperprior model~\cite{balle2018variational} captures the global information in $y$ using the hyper analyzer $h_a$, and the resulting $\hat{z}$ is transmitted to the decoder as side information. The context model~\cite{minnen2018joint, lee2018contextadaptive, minnen2020channel} aims to further save bits by exploiting the correlation between already decoded symbols $\hat{y}_{<i}$ and the currently decoding symbol $\hat{y}_i$. By jointly combining these two methods together, the entropy parameter of each symbol can be formulated as:
\begin{equation}
    \Phi_i = (\mu_i,\sigma_i) = g_{ep}(h_s(\hat z), g_{cm}(\hat y_{<i})),
\end{equation}
where $h_s(\hat z)$ denote the hyperprior feature and $g_{cm}(\hat y_{<i})$ denote the context feature. The two features are fused by $g_{ep}$ before yield probability parameter $\Phi_i$.

\subsection{Multi-generation Robust Coding}
Multi-generation coding is a process of repeatedly compressing decoded pictures~\cite{RobustMPEG}, and the resulting generation loss will have a non-negligible impact on multimedia applications. For traditional coding standards like JPEG and BPG (i.e., HEVC Intra Coding), prior works have shown that inconsistent mode decisions, rounding and clipping (RC) errors are sources of generation loss~\cite{joshi2000comparison, RobustHEVC}. Some work ~\cite{hurd2000achieving} devoted to eliminating the generation loss and achieving idempotent encoding in JPEG-LS, but additional bit allocation is needed.

Multi-generation robustness for LIC was first discussed in previous work~\cite{2020Instability}. A SIC benchmark was conducted for the state-of-the-art learning-based models, results show that most models suffer from serious quality loss during repeated compression. Following their work, the definition of SIC is as follows:
\begin{align}
    \label{equation:SIC}
    f_{SIC} &= RC \circ g_s \circ Q \circ g_a, \\
    x_n &= f_{SIC}^n(x_0),
\end{align}
where $f_{SIC}(\cdot)$ denotes one compression-decompression cycle, following a pipeline of analysis transform $g_a$, quantization $Q$, synthesis transform $g_s$, rounding and clipping $RC$. $x_0$ is the original image, and $x_n$ is the image after $n$ times SIC cycles.

To reduce the multi-generation loss, feature identity (FI) loss~\cite{2020Instability} is proposed to be added during training time, as illustrated in the equation:
\begin{equation}
    \mathcal{L}_{FI} = || \hat y_1 - \hat y_0 ||_2,
    \label{equation:FI}
\end{equation}
where $\hat y_0 = Q(g_a(x_0))$ and $\hat y_1 = Q(g_a(g_s(\hat y_0)))$. Although the degradation rate of image quality was slowed down by using the loss function, the degradation trend has not changed. The reason may be they did not address the problem of quantization drift, which will be discussed in the next section.

Another most relevant work claims to realize approximately lossless reecoding~\cite{helminger2021lossy}. By leveraging normalizing flows, an invertible transformation from the image space to a latent representation is learned. Through this special property, the model can attain approximately lossless successive compression, which is similar to JPEG. However, restricted by the capabilities of the additive coupling layers, the model shows a pool rate-distortion performance at low bitrate (roughly on par with JPEG2000). We will show that an invertible network with a more capable transformation may result in more severe generation loss. Nonetheless, the main focus of our paper is to improve the robustness of the more widely used variational image compression frameworks.

\subsection{Adversarial Attacks on Neural Compression}
There are some works on attacks and defenses for LIC~\cite{liu2022denial, chen2021towards}. They found that adding specific noise to the original image can impact the performance of the LIC. Researchers have proposed several methods to defend against the problem, enhancing the generalization ability of the model. The difference with our work is that they focus on the first encoding performance for different images, we focus on the robustness of the same natural image in the continuous compression process.

\begin{figure}[t]
\centering
\includegraphics[width=1\linewidth]{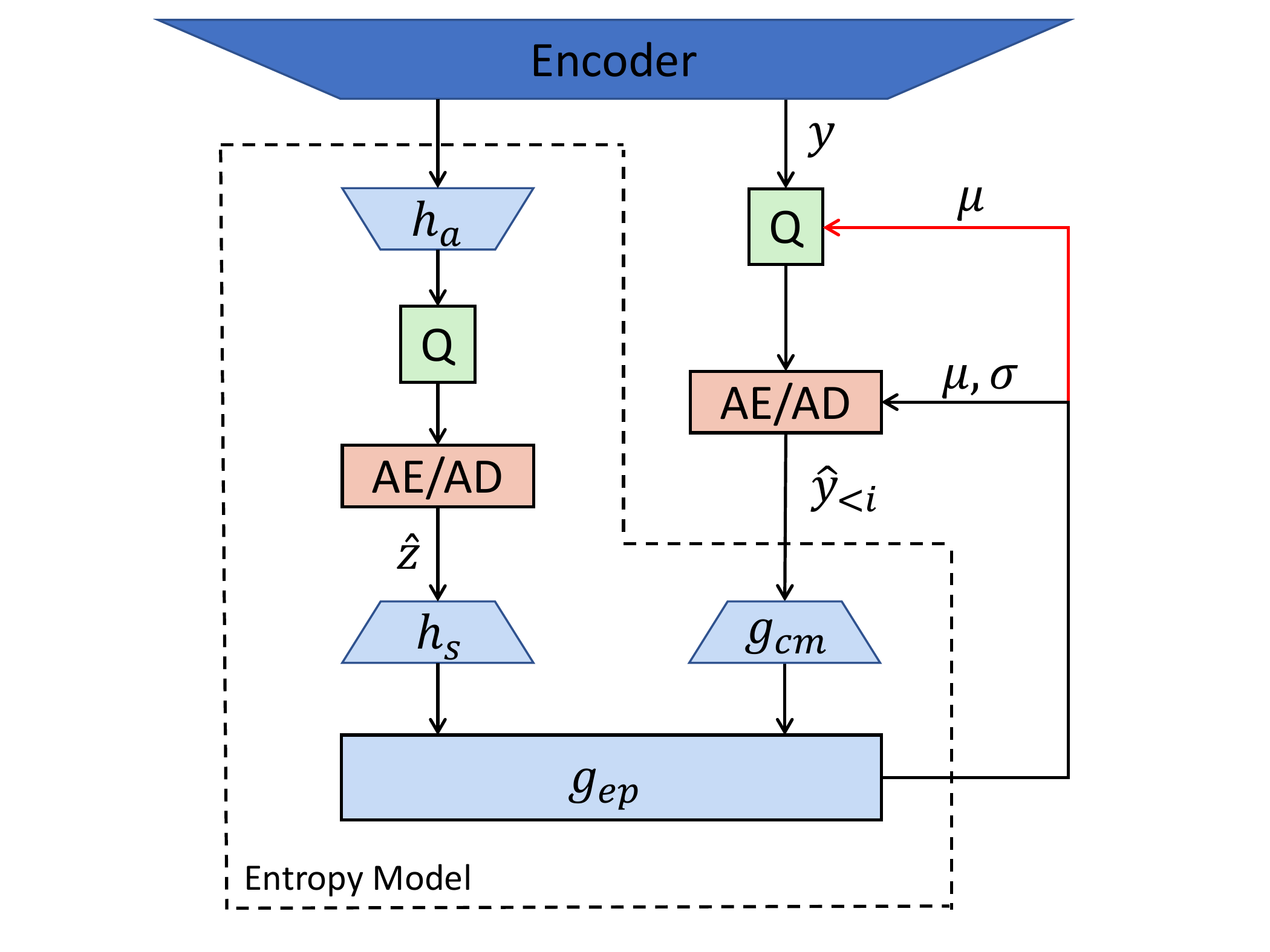}
\caption{Entropy model and quantization implementation. Corrected quantization will subtract $\mu$ computed by the entropy model before quantization, as shown by the red line.}
\label{Figure2}
\end{figure}

\begin{figure*}[t]
    \includegraphics[width=1\linewidth]{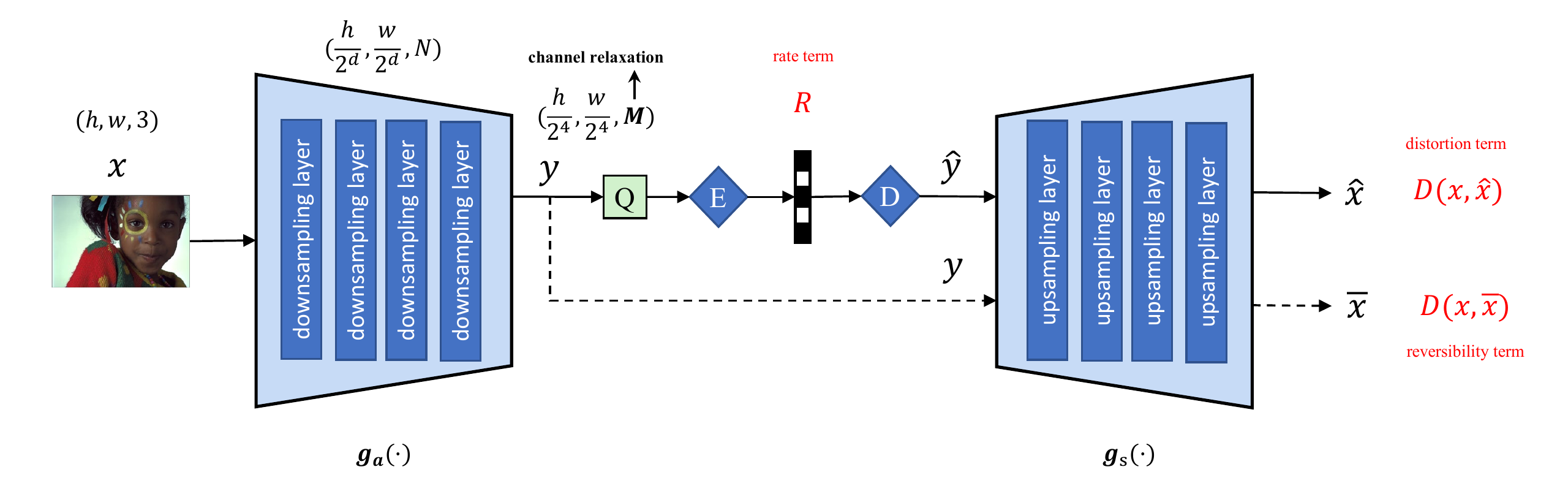}
    \caption{Illustration of basic variational compression framework  and the proposed reversibility loss.}
    \label{Figure3}
\end{figure*}

\section{What Impacts Robustness}\label{sec:analysis}
Before illustrating our methods, we thoroughly analyzed the influence of each component of SIC on multi-generation robustness, as shown in Equation~\ref{equation:SIC}.

\subsection{Quantization Drift Problem}
Quantization is an important module of image compression, and its implementation is crucial to the stability of SIC. In the ideal case of idempotent reencoding, the quantized latent representation $\hat y_0$ and $\hat y_1$ should be equal in two-round encoding, so that the decoded images are also the same. This is the inspiration for the feature identity loss~\cite{2020Instability} in Equation~\ref{equation:FI}, though the RC operations in the decompression-compression-cycle are not considered due to the non-differentiable problem during training.

Recent work tends to follow the quantization implementation in~\cite{balle2020nonlinear}. When encoding symbol $y_i$, they first subtract its mean $\mu_i$, followed by quantization and entropy encoding, and finally add $\mu_i$. The whole quantization process can be expressed by the following formula, which we call ``corrected quantization'':
\begin{equation}
    \hat y_i = round(y_i-\mu_i) + \mu_i,
    \label{corrected quantization}
\end{equation}
where the $\mu_i$ is from the entropy model.

Corrected quantization is widely used in the state-of-the-art works~\cite{balle2018variational, minnen2018joint, minnen2020channel, xie2021enhanced}, and it is also implemented as the default method in the Tensorflow Compression~\cite{tfc} and CompressAI~\cite{begaint2020compressai} platforms. However, we point out that corrected quantization can seriously impact the multi-generation robustness of LIC. As illustrated in Figure~\ref{Figure2}, in an entropy model with hyperprior and context, the output $\mu$ depends on the feature $y$ and the already decoded symbol $\hat y_{<i}$. Assume that $y$ is slightly perturbed, $\mu$ will change and lead change in $\hat y$, and a chain reaction will further occur due to the context model. In this case, to make $\hat y_0$ and $\hat y_1$ equal in two-round encoding, the latent representation $y_0$ and $y_1$ must be the same. However, this is almost impossible due to the existence of quantization and clip errors in the SIC cycle. ``mbt18'' and ``mbt18+FI'' in Figure~\ref{Figure1} shows that corrected quantization causes perturbations to accumulate during successive compression, resulting in continuous degradation of image quality.

\subsection{Transform Reversibility}
The reversibility of the transformation directly affects the quality loss during successive compression. JPEG is based on a transform coding framework. The image is modeled as fixed-size blocks of $8 \times 8$ pixels, and then discrete cosine transform (DCT), scalar quantization, and inverse discrete cosine transform (IDCT) are applied to each block separately. The generation loss of JPEG is trivial due to the manually designed invertible orthogonal transformation.

Existing variational compression models are also based on transform coding~\cite{Kingma2014}, but different from JPEG, they rely on neural networks to construct nonlinear transform $g_a(\cdot)$ and $g_s(\cdot)$. The transformation may consist of architectures such as convolution networks, residual blocks~\cite{he2016deep} and attention modules~\cite{vaswani2017attention}. As shown in Figure~\ref{Figure3}, $g_a(\cdot)$ generally contains 4 downsampling operations, the feature map of the middle layer contains $N$ channels, while latent representation contains $M$ channels. A symmetric architecture is usually adopted by $g_s(\cdot)$.

The network parameters are optimized for rate-distortion function, where the distortion term is $d(x, g_s(\hat y))$ and $\hat y$ is the quantized approximation of $g_a(x)$. The loss function does not explicitly constrain the reversibility of $g_a(\cdot)$ and $g_s(\cdot)$. And to obtain a more compact latent representation, limited $M$ is set in previous works~\cite{balle2018variational, minnen2018joint, cheng2020learned}, which may also restrict the reversibility of the transform network.

\subsection{Model Adaptability}
Another factor that affects multi-generation robustness is the adaptive ability of the model. The researchers showed that error caused by RC operations has little effect on JPEG~\cite{joshi2000comparison}. For LIC that employ nonlinear transformations, the ability to accommodate this error is critical, especially since there is no similarly distorted image in the training set. The invertible network with linear bijective mapping satisfies reversibility, which can theoretically achieve idempotent reencoding if ignore the RC error. Our toy experiment will demonstrate that the adaptability to RC error is important, and the invertible networks with stronger transformation capabilities can be more vulnerable.

\section{Proposed Method}\label{sec:method}
In this section, we first illustrate the proposed scheme to address the quantization drift problem, then describe two different approaches to enhance the reversibility of existing variational compression models.

\subsection{Straight Quantization}
To address the quantization drift problem, we propose to use straight quantization instead of corrected quantization at testing time. In the process of straight quantization, rounding and entropy coding are performed directly on $y_i$ instead of $y_i - \mu_i$, using the following quantization process instead of the Equation~(\ref{corrected quantization}):
\begin{equation}
    \hat y_i = round(y_i).
    \label{equation:straight quantization}
\end{equation}

The latent representation $\hat y$ fed into the decoder is a discrete value, and the quantization process of each symbol $y_i$ is relatively independent, no longer depends on the whole $y$ and the already decoded $\hat y_{<i}$. In the process of successive compression, as long as the values of $y_0$ and $y_1$ fall in the same quantization interval, the quantized value $\hat y_0$ and $\hat y_1$ will be the same and so does the decoded images.

To avoid train-test mismatch, we use straight-through estimator (STE)~\cite{TheisSCH17} rather than noisy approximation for training. STE applies hard rounding in the forward pass and uses the modified gradient in the backward pass. This is equivalent to optimizing a deterministic autoencoder~\cite{guo2021soft}, which we find effective for multi-generation robust coding. As for entropy rate estimation, we still follow the mainstream works and adopt the method of adding uniform noise.

\subsection{Reversibility Loss Function}
There is no explicit constraint on the reversibility of the transformation in the original loss function (Equation~\ref{equation:RD}). To enhance the reversibility and reduce the generation loss, we add a reversible constraint to the original distortion term at training time, the new loss function can be written as:
\begin{equation}
    \mathcal{L} = R(\hat y) + \lambda (D(x, \hat x) + \alpha D(x, \overline x)),
    \label{equation:reversibility loss}
\end{equation}
where $\overline x = g_s(y)$ and $\alpha$ is a hyperparameter to control a trade-off between real distortion term and reversible constraint term. The approach is shown in Figure~\ref{Figure3}. Different from feature identity loss proposed in ~\cite{2020Instability}, reversibility loss does not consider the quantization process. We directly send the latent $y$ before quantization into the decoder and obtain the image $\overline x$, and expect $\overline x$ to be equal with the original image $x$, thus constraining the reversibility of $g_a$ and $g_s$. Empirically, our approach will make the final model more stable during SIC.


\subsection{Channel Relaxation}
Existing LIC models use a transformation network $g_a(\cdot)$ with 4 spatial downsampling operations to transform the original image $x \in \mathbb{R}^{3 \times h \times w}$ into a compact representation $y \in \mathbb{R}^{M \times \frac{h}{16} \times \frac{w}{16}}$, where $h$ and $w$ denote the height and width of the original image and $M$ denotes the channel number of latent feature $y$. In previous works~\cite{balle2018variational, minnen2018joint, cheng2020learned}, $M$ is generally set to be larger in high bitrate models. The reason may be that the latent with a larger dimension can retain more information at the entropy bottleneck, thus enabling the decoder to recover more details.

Our pre-experiments confirm that limited channels constrain the information-holding ability of the latent representation. Too much information is lost in transformation, resulting in weaker reversibility of the transform network. Dimensional relaxation of the compact latent $y$ can reduce the constraints, making information loss in the quantization process instead of transformation, which coincides with traditional codecs. The method will reduce the generation loss without changing the network architecture and optimization function. To implement channel relaxation in existing models, we obtain hyperparameter $M$ suitable for every bitrate through experiments, and $M$ will affect the parameter dimensions of the layers close to the entropy bottleneck (eg. input and output channel numbers of convolutional layers).

\section{Experiments}\label{sec:experimentalresult}
In this section, we first prove the superiority of the straight quantization strategy, followed by detailed testing and comparison of two different solutions for enhancing robustness. Furthermore, the significance of model adaptability is explored through the analysis of invertible networks.

\subsection{Experiment Setup}
\subsubsection{Baseline Models} We choose the joint autoregressive and hierarchical entropy model ``mbt18''~\cite{minnen2018joint} as well as its two variants ``mbt18mean'' and ``cheng20anchor''~\cite{cheng2020learned} as baseline models. Different from mbt18, mbt18mean only uses hyper analyzer for probability estimation, and cheng20anchor uses stacked residual blocks for analysis and synthesis networks instead of ordinary convolutional layers in mbt18. When evaluating the baseline models, we use the pre-trained weights in CompressAI~\cite{begaint2020compressai} corresponding to six reconstruction qualities ranging from quality = 1 to 6.

\subsubsection{Training details} We use Flicker 2W dataset used in~\cite{flicker} for training, which consisting of 20, 745 high-quality images. The images are cropped as $256 \times 256$ patches before being input into the networks. All the models are trained for 1.8M steps with a batch size of $8$ using Adam optimizer, with an initial learning rate of $10^{-4}$ and reduced to $10^{-5}$ for the last 0.2M steps. Models are optimized with MSE (mean square error) quality metric, which is consistent with traditional codecs. $\lambda$ is chosen from the set \{0.0016, 0.0032, 0.0075, 0.015, 0.03, 0.045\}.

\subsubsection{Evaluation details} We evaluate learned models and traditional codecs on three commonly used datasets for image compression, which are Kodak dataset~\cite{kodak}, the CLIC Professional dataset~\cite{CLIC2020}, and the Tecnick dataset~\cite{asuni2014testimages}. Consistent with previous work~\cite{2020Instability}, we set the number of SIC cycles to $n = 50$. We not only evaluate the loss of PSNR during successive compression, but also the rate-distortion performance at $n$-th time, where the distortion is computed by $D(x_0, x_n)$, and the rate is $R(x_{n-1})$.

\begin{figure}[htbp]
    \centering
    \includegraphics[width=1\linewidth]{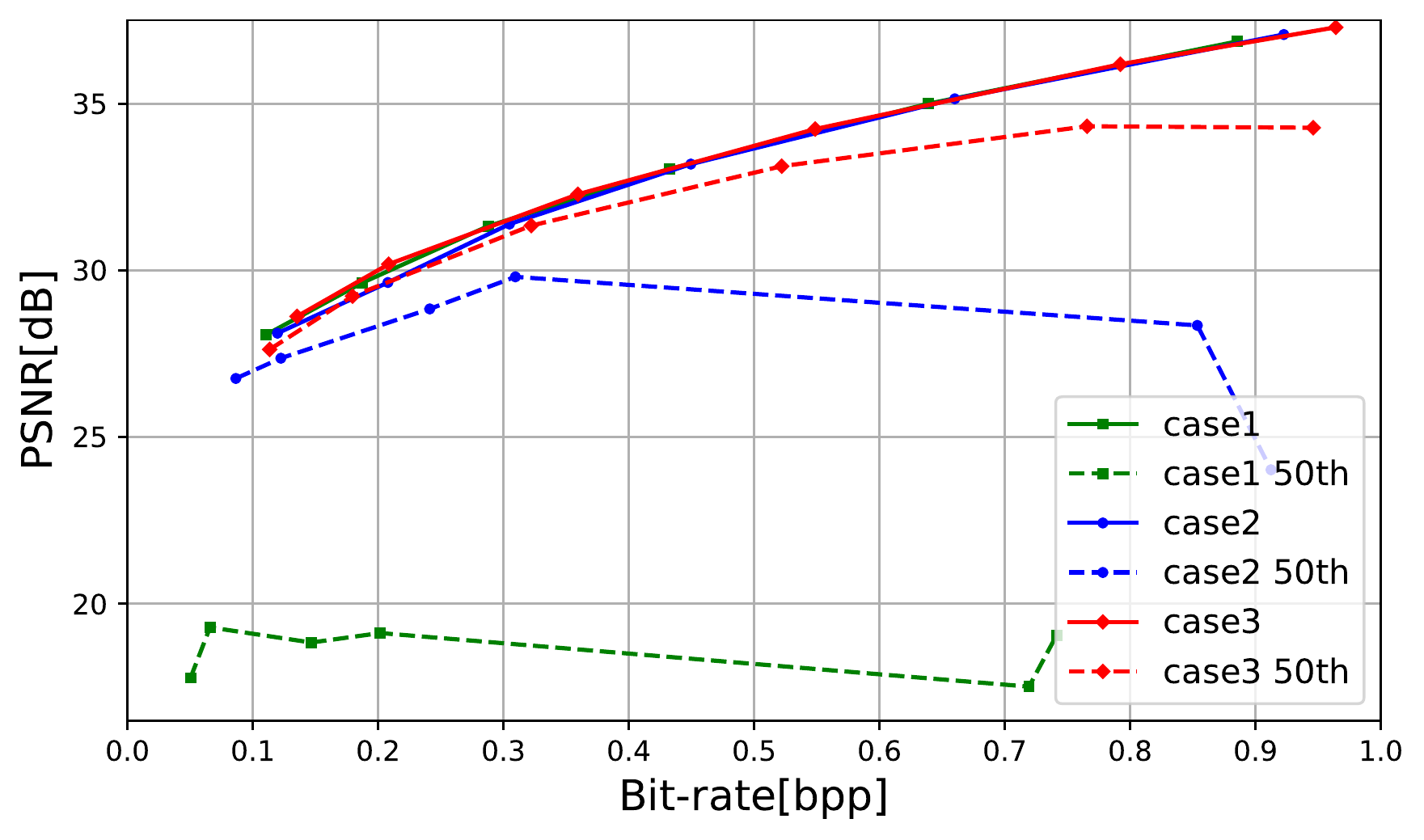}
    \caption{Evaluation of quantization strategies on Kodak.}
    \label{Figure4}
\end{figure}

\begin{table}[htbp]
\renewcommand\arraystretch{1}
\caption{Different quantization strategies.}
\label{tab:quantization}
\centering
\begin{tabular}{|c|c|c|}
\hline
                & \textbf{Training} & \textbf{Testing}       \\ \hline
\textbf{case 1} & uniform noise     & corrected quantization \\
\textbf{case 2} & uniform noise     & straight quantization  \\
\textbf{case 3} & STE               & straight quantization  \\ \hline
\end{tabular}
\end{table}

\subsection{Quantization Strategy}
Quantization strategies include approximate approaches at training time and quantization implementations at testing time, which play a crucial role in multi-generation robustness. We tried the combinations in Table~\ref{tab:quantization}, and the results on the Kodak dataset are shown in Figure~\ref{Figure4}. The rate-distortion performance of the first compression has little difference in the three cases, but the difference becomes significantly larger after repeated compression for 50 times. The same pre-trained models but different quantization implementations are used in case 1 and case 2. It can be seen that the quality loss is significantly reduced when the corrected quantization is replaced by straight quantization. In case 3, the train-test mismatch is eliminated after changing the training approach from noise to STE, resulting in better robustness. We use ``SQ'' to represent the quantization strategy of case 3 in the subsequent experiments.

\begin{figure}[htbp]
    \centering
    \includegraphics[width=1\linewidth]{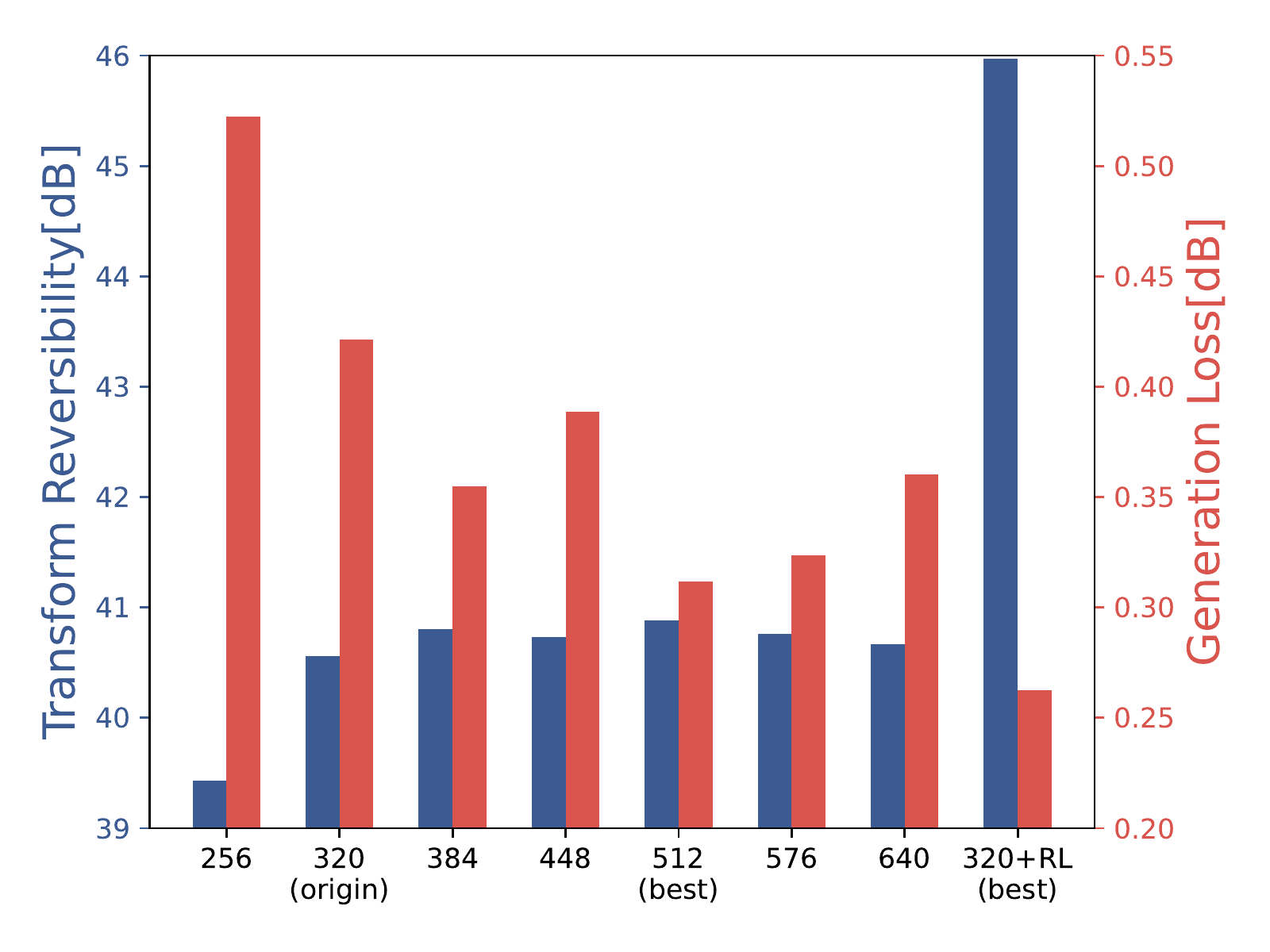}
    \caption{Effect of channel relaxation and reversibility loss. The numbers on the x-axis represent $M$, ``320+RL'' denotes $M=320$ with reversibility loss.}
    \label{Figure5}
\end{figure}

\begin{table}[htbp]
\centering
\renewcommand\arraystretch{1.2}
\caption{Proposed relaxed $M$ for mbt18.}
\label{tab:channels}
\begin{tabular}{ccccccc}
\toprule
\textbf{Quality} & \textbf{1, 2, 3} & \textbf{4} & \textbf{5} & \textbf{6} \\ \hline
\textbf{Original Channel} & 192 & 192 & 320 & 320 \\ \hline
\textbf{Channel Relaxation} & 192 & \textbf{448} & \textbf{512} & \textbf{576} \\ 
\bottomrule
\end{tabular}
\end{table}

\subsection{Reversibility Enhancement}
Figure~\ref{Figure4} shows that only using the straight quantization strategy, the generation loss is still large at high bit rates. According to our analysis, channel relaxation (``CR'' for short) or reversibility loss (``RL'' for short) can further improve the robustness of the model. To explore the effect of latent channel number $M$ and the reversibility loss function, $M$ is increased from $256$ to $640$ by $64$ each time, $\alpha$ in Equation~\ref{equation:reversibility loss} is setting to $1$ empirically. We use $d(x_0, \overline{x}_0)$ to characterize transform reversibility. To explore the impact of reversibility on generation robustness, we compute the first-generation loss as $d(x_2, x_0) - d(x_1, x_0)$. The experimental results on the mbt18 model improved by ``SQ'' with quality = 5 are shown in Figure~\ref{Figure5}.

The results reveal that stronger reversibility will lead to smaller generation loss. As the number of channels $M$ increases, the reversibility shows a trend of first increasing and then decreasing. The reason is that slightly increased $M$ relaxes the information bottleneck, and reduces information loss in the transformation. While too large $M$ will lead to unstable and insufficient training. Similar results for different bitrates and models are shown in the supplementary material. The reversibility increases significantly after adding explicit constraints in loss function. Table~\ref{tab:channels} presents the $M$ with the best robustness after channel relaxation for mbt18. Note that only the models with high bitrate are relaxed.

\begin{figure*}[htbp]
    \centering
    \includegraphics[width=0.32\linewidth]{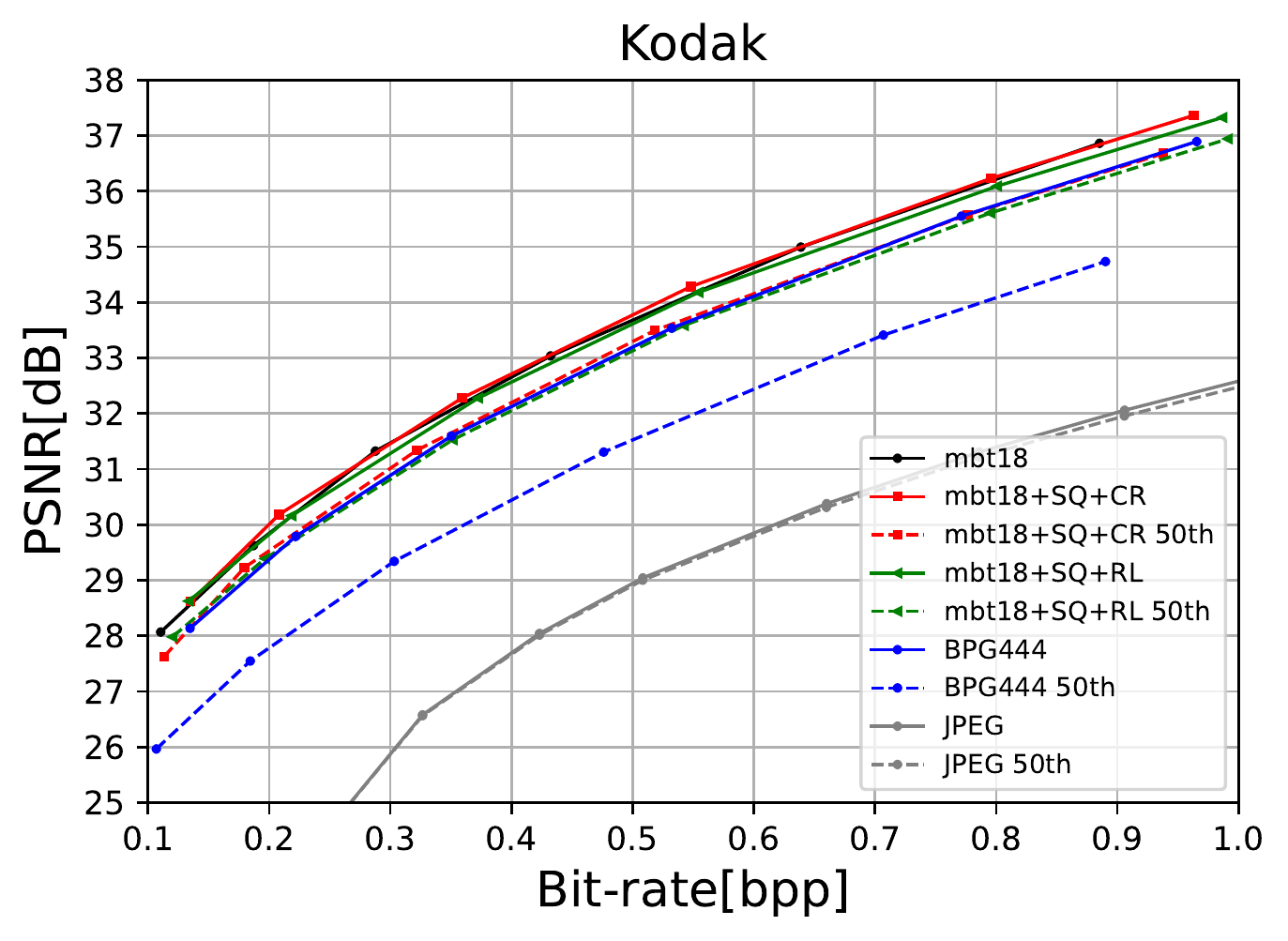}
    \includegraphics[width=0.32\linewidth]{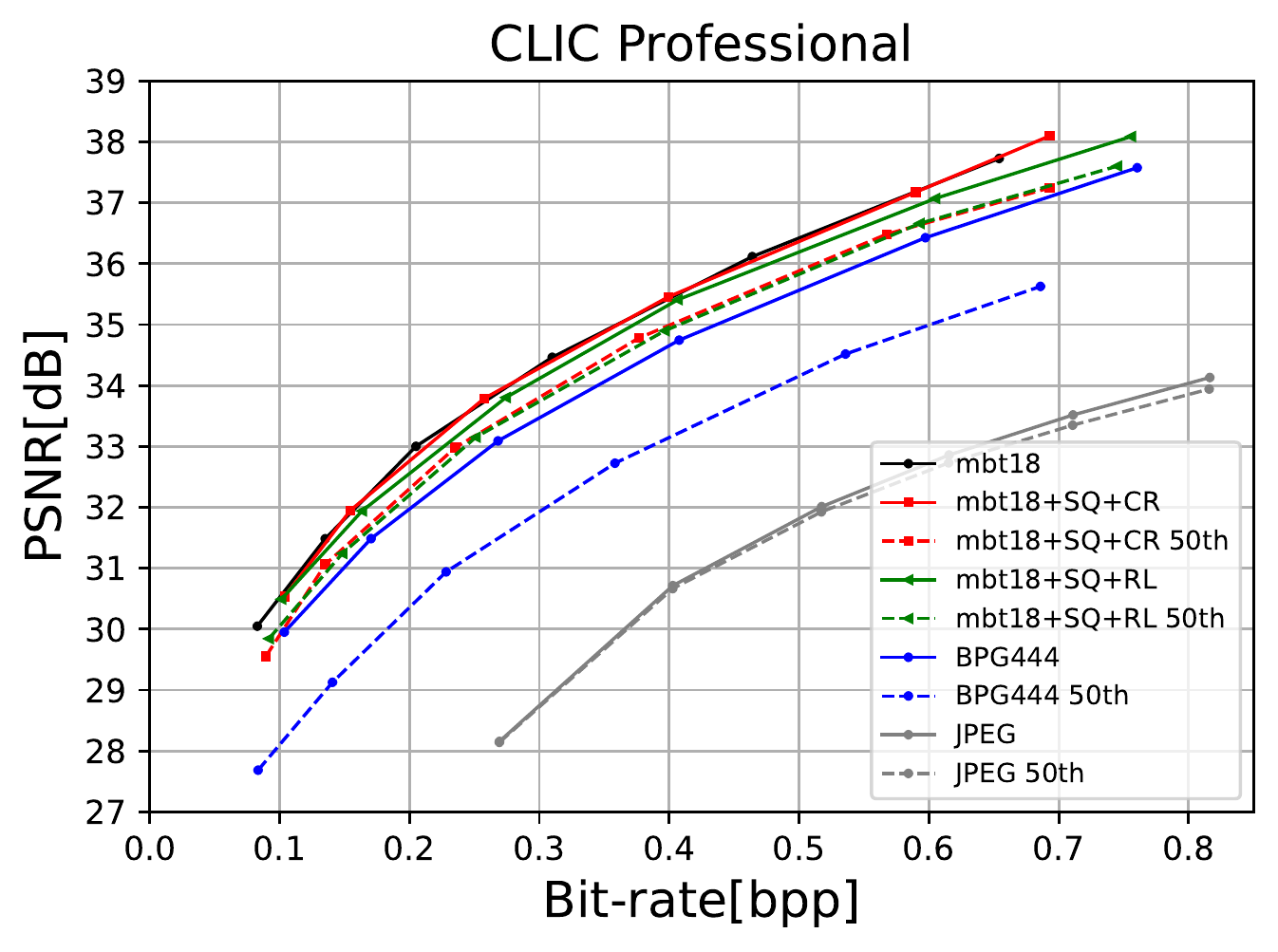}
    \includegraphics[width=0.32\linewidth]{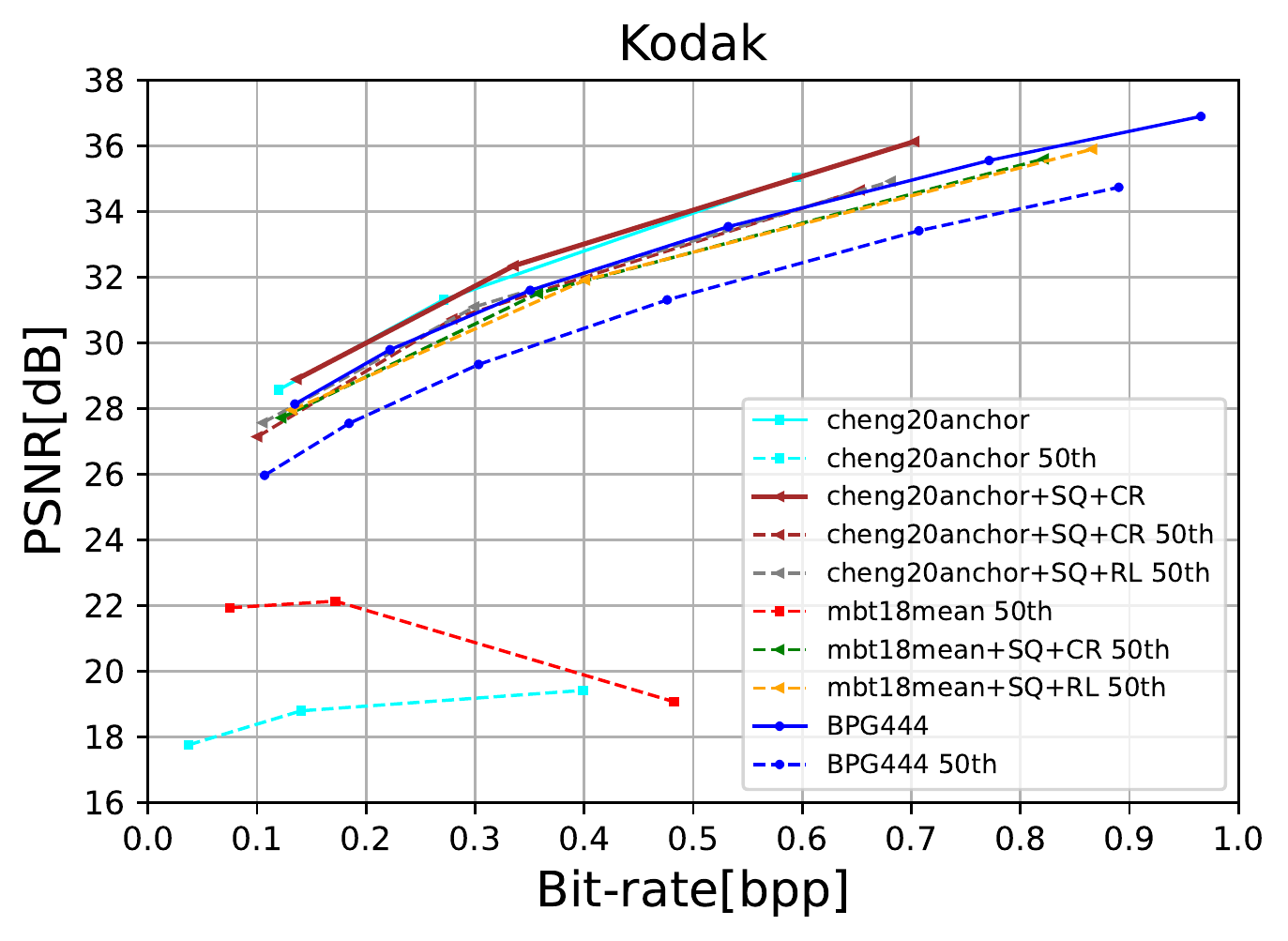}
    \\
    \includegraphics[width=0.32\linewidth]{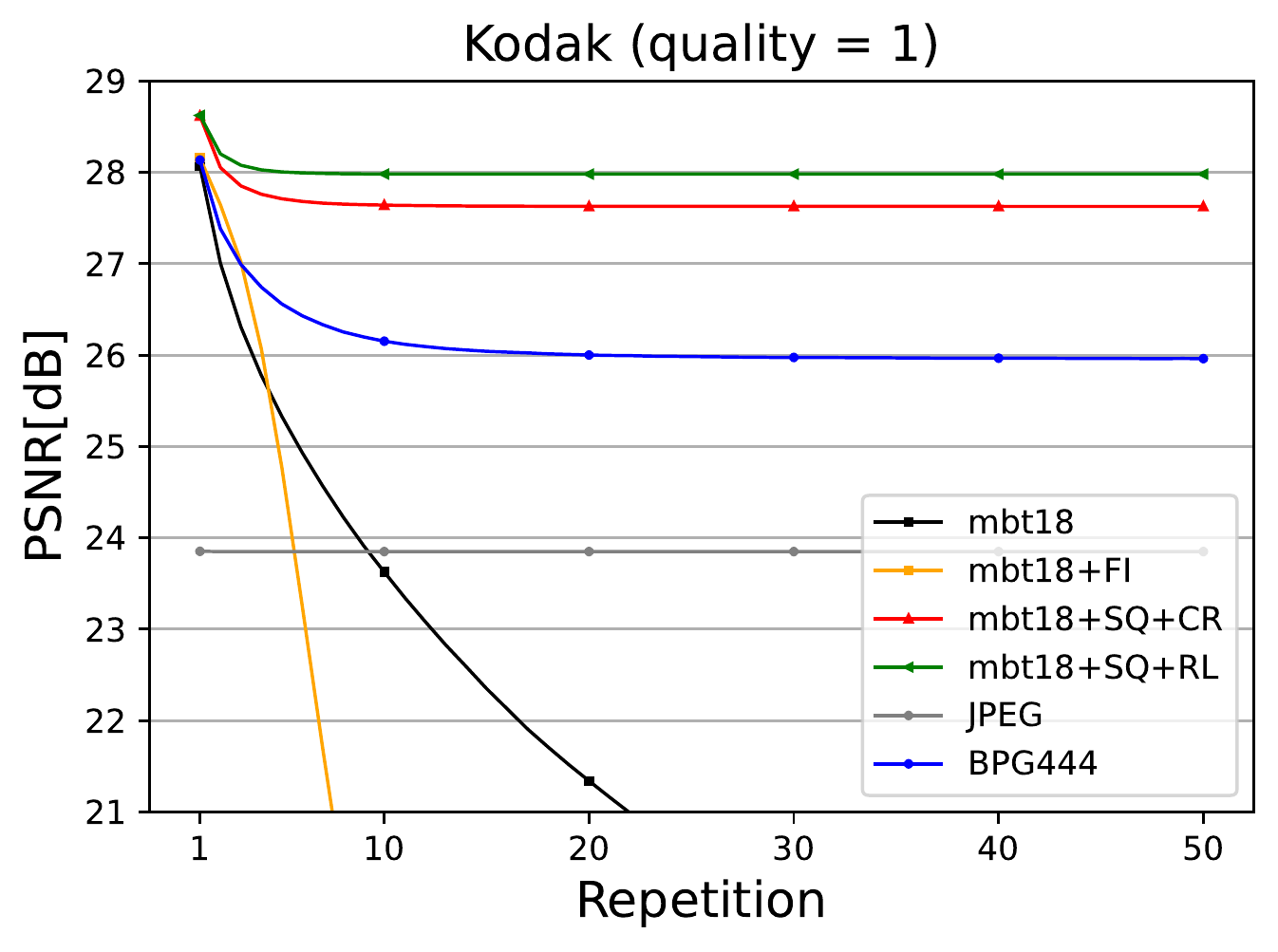}
    \includegraphics[width=0.32\linewidth]{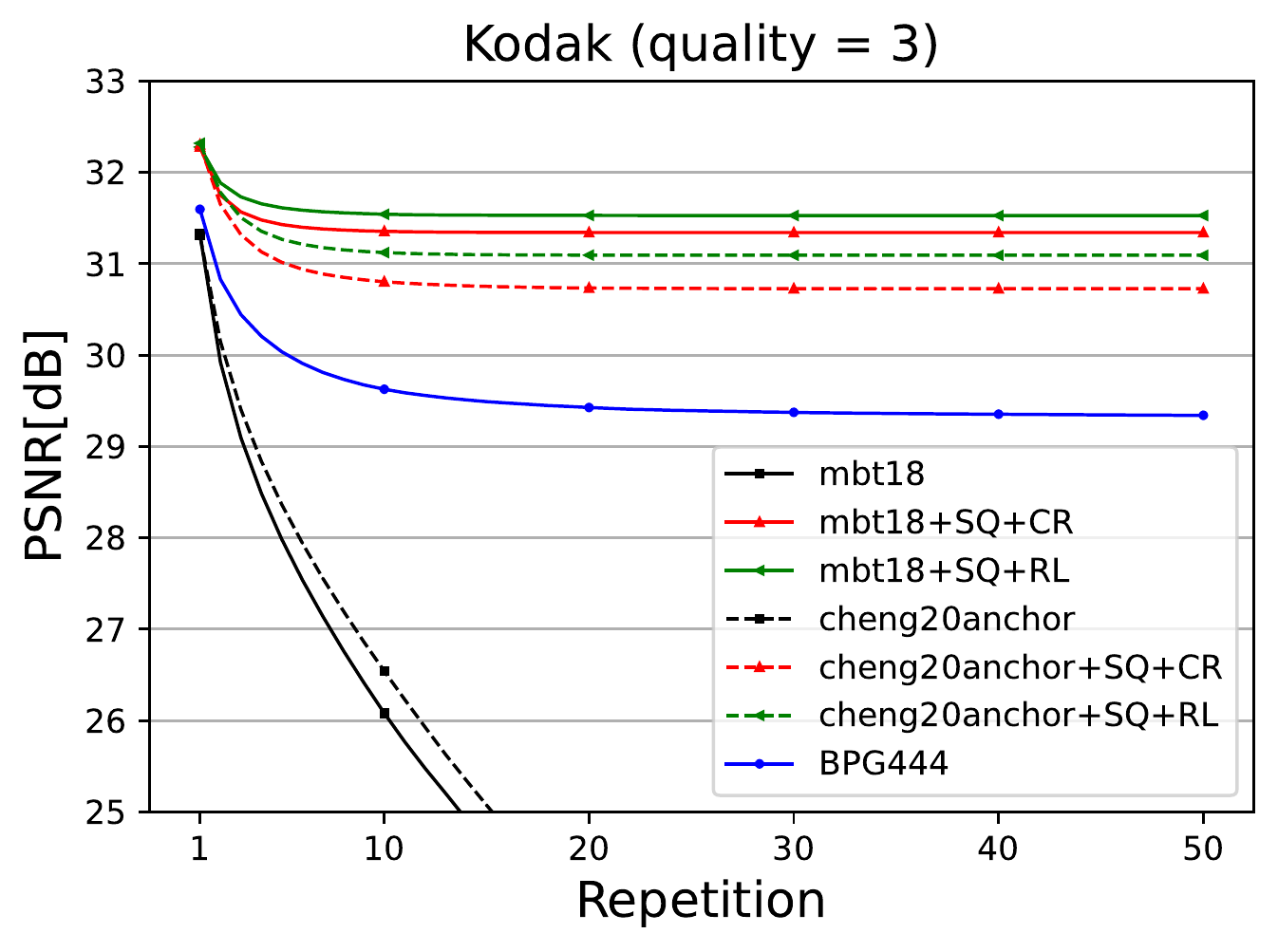}
    \includegraphics[width=0.32\linewidth]{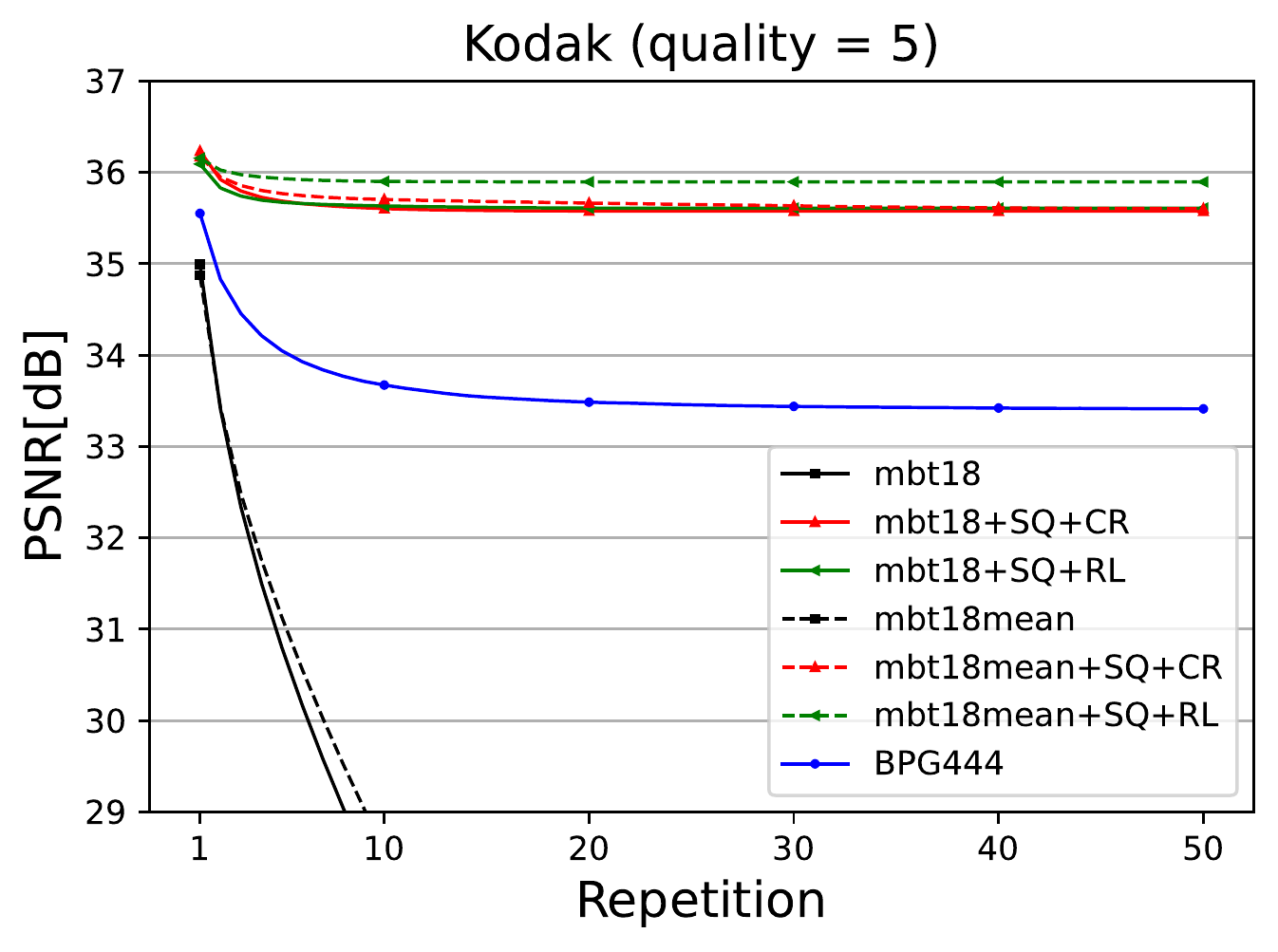}
    \caption{Quantitative evaluation results. Models based on mbt18mean and cheng20anchor are evaluated with quality = 1, 3, 5. Some results are not presented in the figure for clarity, and the complete results are presented in the supplemental material.}
    \label{Figure6}
\end{figure*}

\subsection{Overall Evaluation}
Through pre-experiments, we finally got two solutions to reduce the generation loss, which are ``SQ+CR'' and ``SQ+RL''. The quantitative evaluation results on different datasets and different baseline models are shown in Figure~\ref{Figure6}. During the process of SIC, the image quality degradation quickly converges by using our improved mbt18 model, the average PSNR reduction is less than $1.0$ dB on both datasets. After 50 times reencoding, we can still maintain excellent rate-distortion performance, even comparable to the first encoding of BPG. Experiments on mbt18mean and cheng20anchor also demonstrate the effectiveness of our solutions. More experimental details and results are provided in the supplementary material.

We note that both solutions have pros and cons. ``SQ+CR'' shows better performance in rate-distortion performance, but channel relaxation increases the model parameters at high bitrates. Without changing original settings, ``SQ+RL'' can stop quality degradation faster during SIC, but the rate-distortion performance is slightly impacted due to reversibility term, although it can be flexibly tuned by adjusting $\alpha$.

\begin{figure}[htbp]
    \centering
    \includegraphics[width=1\linewidth]{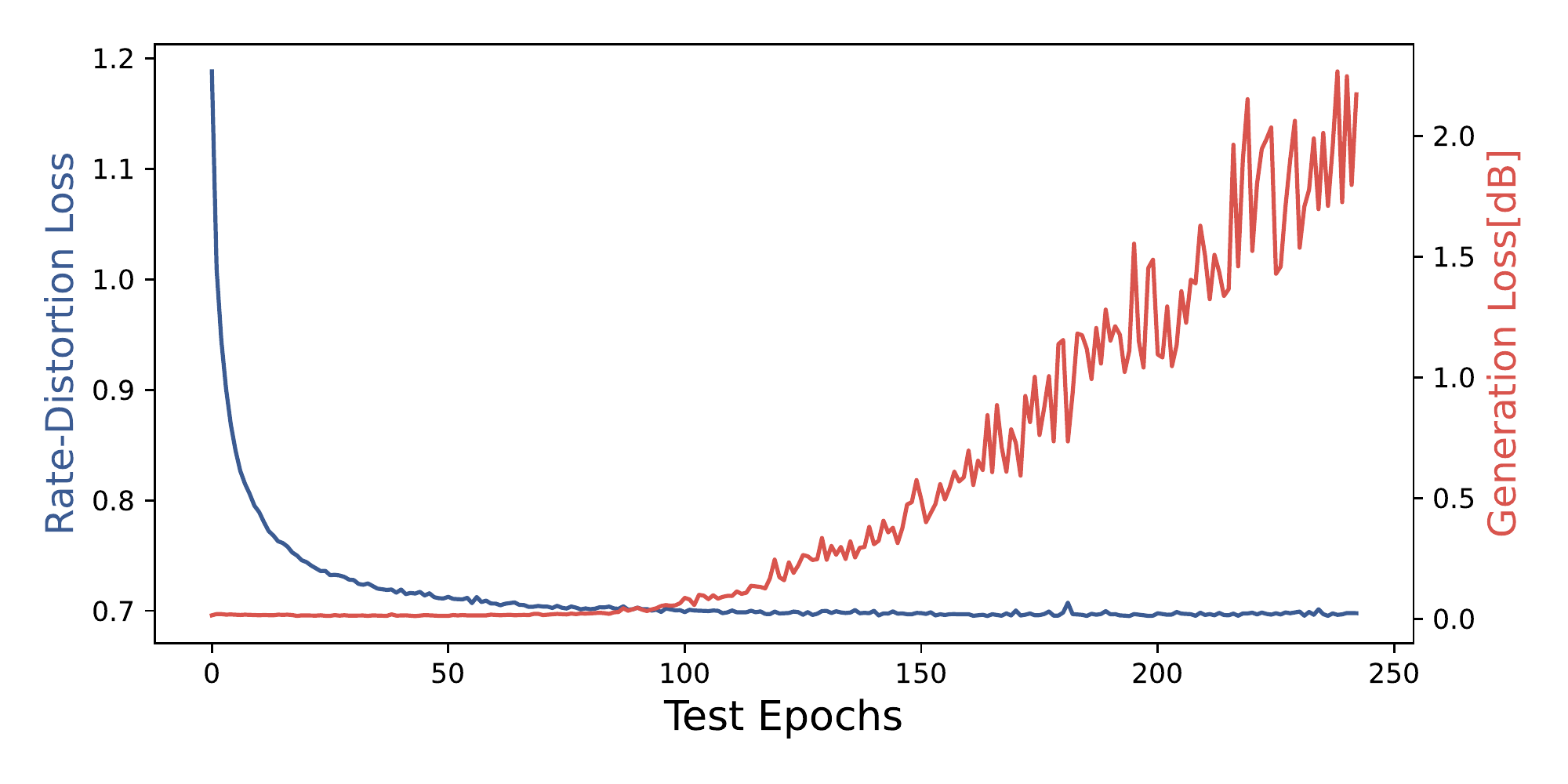}
    \caption{Toy experiment on ``INN+SQ''.}
    \label{Figure7}
\end{figure}

\subsection{Model Adaptability Analysis}

\subsubsection{Toy Experiments on INN} Invertible neural networks (INNs) maintain a bijective mapping between inputs and outputs, which is mathematically reversible. We show through toy experiments that INNs may not suitable for the SIC task. We use the invertible block proposed by ~\cite{xie2021enhanced} to build the transform network and adopt straight quantization strategy. It can achieve comparable coding performance with mbt18 (more details in supplementary material). Figure~\ref{Figure7} shows how the first-generation loss changes on the validation set during training. The coding performance improves while the generation loss becomes larger, which indicates that the INN with stronger transformation ability is vulnerable during SIC. The fragility of INNs is also mentioned in image rescaling cycles~\cite{pan2022towards}.

\subsubsection{Influence of RC errors} 
\begin{figure}[htpb]
    \centering
    \includegraphics[width=1\linewidth]{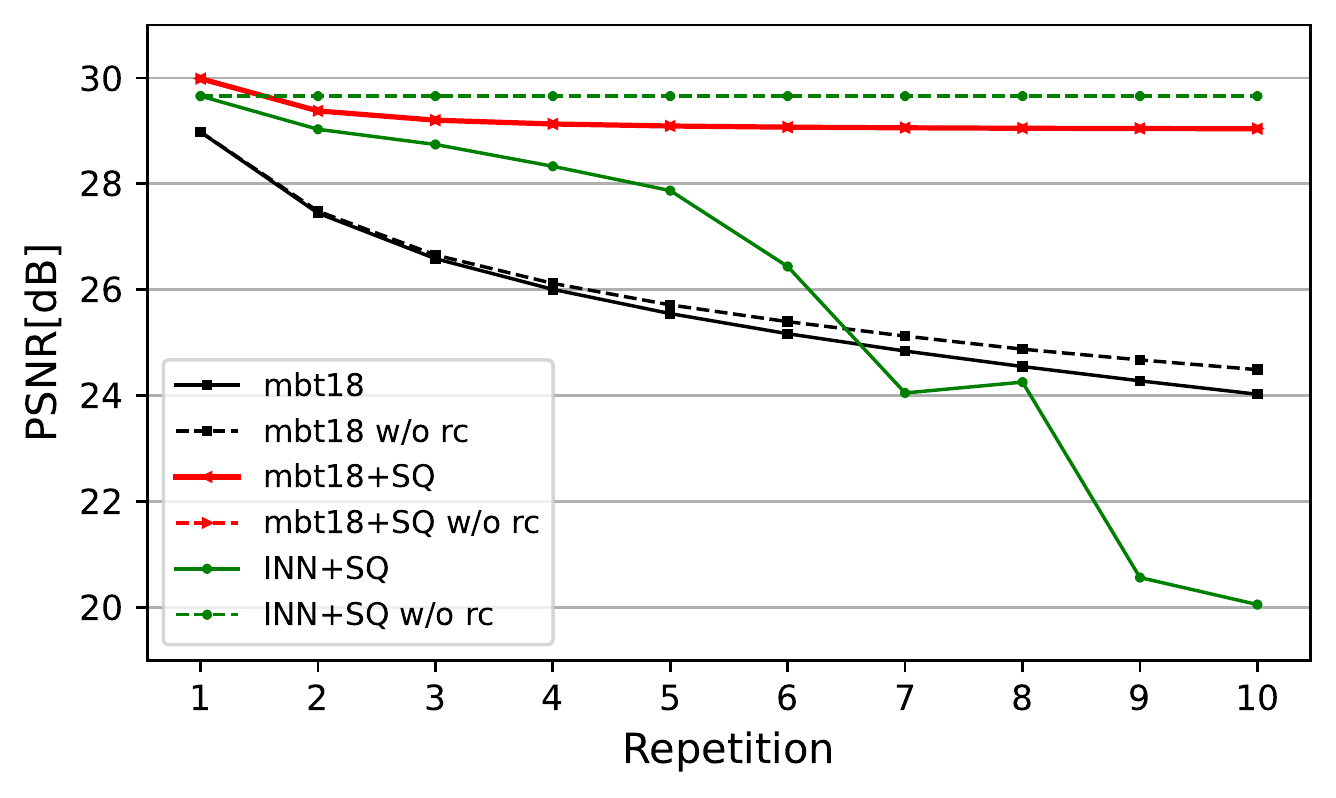}
    \caption{SIC of ``kodim01.png'' with $n=10$. RC errors have an ignorable influence on mbt18+SQ.}
    \label{Figure8}
\end{figure}

In order to verify the influence of RC errors in SIC, we compared the generation loss of mbt18, mbt18+SQ and INN+SQ with and without RC operations, respectively. All the models are trained with quality = 3, results are shown in Figure~\ref{Figure8}. We can see that INN+SQ can achieve idempotent reencoding when ignore RC operations. But in practical situations, the RC error and the quantization process will cause the image quality to degrade rapidly, forming a phenomenon similar to the quantization drift in mbt18. We note that the variational compression frameworks can better at adapting to such errors than INNs, the reason may be that the transformation network acts as a low-frequency filter in the variational models, thereby filtering out the ``unfriendly'' noise in reencoding, while INNs allow the error to be passed between the image domain and the latent domain, thereby accumulating continuously.

\section{Conclusion}\label{sec:conclusion}
In this paper, we thoroughly analyze the factors that affect the multi-generation robustness of LIC. We discovered and solved the quantization drift problem in existing models, and proposed two solutions to further reduce the generation loss. Extensive experiments show that our solutions are effective for different datasets and models. By using the model improved with our methods, after successively compressing the images 50 times, the average PSNR reduction is less than 1.0 dB, while the rate-distortion performance is comparable to the first compression of BPG. The results are acceptable in practical applications, making the future of LIC more promising. In addition, we demonstrate through toy experiments that the variational compression framework has better adaptability to SIC task, while INNs with comparable encoding performance are more susceptible to RC errors.

\section*{Acknowledgements}\label{sec:acknowledgment}
This work is supported by National Natural Science Foundation of China U21B2012, 62072013 and 61902008, Shenzhen Cultivation of Excellent Scientific and Technological Innovation Talents RCJC20200714114435057, Shenzhen Research Projects of JCYJ20180503182128089 and 201806080921419290, Shenzhen Fundamental Research Program (GXWD20201231165807007-20200806163656003). 

\bibliography{aaai22}
\end{document}